%% file: main.tex
\documentclass[runningheads]{llncs}

\usepackage{eccv}

\usepackage{eccvabbrv}

\usepackage{graphicx}
\usepackage{booktabs}

\usepackage[accsupp]{axessibility}  

\usepackage{hyperref}

\usepackage{orcidlink}

\begin{document}

\title{FaceOracle: Chat with a Face Image Oracle}


\author{Wassim Kabbani \and
Kiran Raja \and \\
Raghavendra Ramachandra \and
Christoph Busch}

\authorrunning{W. Kabbani, K. Raja, R. Ramachandra, C. Busch}

\institute{Norwegian University of Science and Technology, Gjøvik, Norway
\email{\{wassim.h.kabbani,kiran.raja,raghavendra.ramachandra,christoph.busch\}@ntnu.no}}

\maketitle

\input{00_abstract}
\input{01_introduction}
\input{02_related_work}
\input{03_approach}
\input{04_experiments}
\input{05_discussion}
\input{06_conclusion}
\input{10_other}

\clearpage  

%
%
\bibliographystyle{splncs04}
\bibliography{main}
\end{document}

%% file: 00_abstract.tex
\begin{abstract}

    A face image is a mandatory part of ID and travel documents. Obtaining high-quality face images when issuing such documents is crucial for both human examiners and automated face recognition systems. In several international standards, face image quality requirements are intricate and defined in detail. Identifying and understanding non-compliance or defects in the submitted face images is crucial for both issuing authorities and applicants. In this work, we introduce FaceOracle, an LLM-powered AI assistant that helps its users analyze a face image in a natural conversational manner using standard compliant algorithms. Leveraging the power of LLMs, users can get explanations of various face image quality concepts as well as interpret the outcome of face image quality assessment (FIQA) algorithms. We implement a proof-of-concept that demonstrates how experts at an issuing authority could integrate FaceOracle into their workflow to analyze, understand, and communicate their decisions more efficiently, resulting in enhanced productivity.
    
    \keywords{FIQA \and Explainable FIQA \and RAG \and LLMs \and AI Assistant}
\end{abstract}

%% file: 01_introduction.tex
\section{Introduction}
\label{sec:intro}

A face image must be included in passports, citizen cards, and residence permits in accordance with the requirements of the International Civil Aviation Organization (ICAO) specifications \cite{ICAO-9303-p1-2021, ICAO-9303-p7-2021}. There are many requirements that a face image needs to meet in order to be compliant. These requirements are outlined in detail in international standards \cite{ISO-IEC-39794-5-G3-FaceImage-191015, ISO-IEC-19794-5-G2-FaceImage-110304}. During the application process for an ID or travel document, and depending on how automated this process is, applicants and issuing authorities need to make sure that submitted face images conform to the quality requirements of both international standards as well as any additional requirements set by the issuing authority. In countries that offer online ID and travel document issuance applications \footnote{https://www.ireland.ie/en/dfa/passports/passport-online/} \footnote{https://www.cbp.gov/travel/us-citizens/mobile-passport-control}, the applicant uploads a photo of themselves as part of the application process before it is forwarded for processing by the issuing authority. Typically, a set of automated checks will process and analyze the face image to determine compliance, quality, and potential manipulations \cite{ISO-IEC-29794-5-DIS-FaceQuality-240129}. Some of these checks could be exposed to the applicant when uploading the image so that they can change the image before submitting the application, but eventually the results of these checks are presented to the case handlers and experts in the issuing authority to assist them in detecting any issues during the processing of the application. Researchers recently conducted a study \cite{Godage-HumanAbilityMAD-TTS-2023} on 469 human observers, including border guards and ID document experts, to assess their proficiency in identifying morphed images. The study revealed that expert human observers performed significantly less accurately than automated morphing attach detection (MAD) algorithms. Yet, it is primarily the responsibility of human experts to make decisions on applications based on their best judgment of the information presented to them. Since many of the new algorithms used to check the compliance, quality, and manipulation of face images are based on AI techniques in general, the idea of explainable AI is needed to help make the results of these algorithms easier to understand \cite{Dwivedixai23, Randyxai18}.

Large Language Models (LLMs) are advanced machine learning models that understand, generate, and manipulate human text \cite{yao2024survey}. These advanced AI systems have made significant strides in natural language understanding and generation, making them ideal for building conversational systems that enhance the user experience and operational efficiency \cite{RaiaanApp2024, yang2024harnessing}. This makes LLMs a suitable candidate for building a conversational interface to a system that can help users analyze and understand the quality of a face image. However, this prospect presents two challenges. First, unlike the generic tasks, such as image captioning or generation, that some LLMs can handle \cite{RaiaanApp2024}, face image quality assessment tasks are very specialized, and general-purpose LLMs cannot perform domain-specific tasks in a trusted way \cite{harvel2024can}. Second, not all of the documents that describe the specifications a face image must have in order to be compliant are publicly available. Some of these documents are standards that are not freely available on the internet or internal policies specific to an issuing authority that define further customized requirements, and thus a general-purpose LLM cannot be expected to have been trained on them \cite{yang2024harnessing}.

In this work, we introduce FaceOracle, an AI assistant specialized in face image analysis, particularly face image quality assessment (FIQA). We use Retrieval Augmented Generation (RAG) to repurpose a pre-trained LLM for FIQA's specialized domain and give FaceOracle access to external knowledge sources such as standards, internal policies, and technical reports. We further use the concept of LLM-powered autonomous agents to give FaceOracle access to the most recent set of FIQA algorithms that are compliant with the ISO/IEC-29794-5 international standard \cite{ISO-IEC-29794-5-DIS-FaceQuality-240129}. This ensures that FaceOracle uses compliant algorithms to perform its computations and provides precise answers based on standards. Experts in an issuing authority can greatly increase their productivity by using FaceOracle, which enables them to perform a comprehensive set of FIQA tasks in a natural and conversational manner, leveraging the power of LLMs, and most importantly, enhancing their ability to interpret, justify, and communicate their decisions. 

%% file: 02_related_work.tex
\section{Related Work}
\label{sec:related-work}

\subsection{Face Image Quality Assessment (FIQA)}

Face Image Quality Assessment (FIQA) refers to the process of determining the utility of a face image to a face recognition system (FRS) \cite{schlettFaceImageQuality2022}. FIQA algorithms can be classified into two main categories: those that assess the image as a whole and those that assess a particular aspect of the image \cite{ISO-IEC-29794-5-DIS-FaceQuality-240129}. Several standards have been proposed to formalize the various aspects that determine the quality of a face image \cite{ISO-IEC-19794-5-G2-FaceImage-110304, ISO-IEC-29794-5-DIS-FaceQuality-240129}. According to the ISO/IEC 29794-5 international standard \cite{ISO-IEC-29794-5-DIS-FaceQuality-240129}, the term \textit{unified quality score} is used to refer to a measure that assesses the image as a whole, and the term \textit{component} is used to refer to a measure that assesses an individual aspect of the image. It further divides the \textit{components} into capture-related components (such as illumination uniformity, the presence of occlusions, and over-/under-exposure) and subject-related components (such as pose, expression neutrality, and mouth closed) \cite{ISO-IEC-29794-5-DIS-FaceQuality-240129}. FIQA algorithms that measure a specific component produce a \textit{quality component value}, and the ones that assess the image as a whole produce a \textit{unified quality score} \cite{ISO-IEC-29794-5-DIS-FaceQuality-240129}. RankIQ \cite{JianshengRankIQ2015}, FaceQnet \cite{HernandezOrtega-FQA-FaceQnetV1-2020}, MagFace \cite{Meng-FRwithFQA-MagFace-CVPR-2021}, and SER-FIQ \cite{terhorstSERFIQUnsupervisedEstimation2020} are examples of deep learning end-to-end methods that produce a unified quality score. NeutrEx \cite{Grimmer-NeutrEx-IJCB-2023} and Syn-YawPitch \cite{grimmerPoseImpactEstimation2023} are examples of methods that target a specific quality component, herein neutral expression and pose, respectively.

\subsection{Explainable Face Image Quality Assessment}

Explainable AI (XAI) refers to methods and techniques in artificial intelligence that make the behavior and outcomes of AI systems more understandable to humans \cite{speith2022review}. The goal of XAI is to create AI models and systems whose actions and decisions can be easily interpreted, communicated, and trusted by humans \cite{Dwivedixai23, Randyxai18}. Most works in the literature that aim at FIQA explainability focus on end-to-end FIQA methods that produce a unified quality score \cite{terhorst2023pixel, fuRelativeContributionsFacial2021, fuExplainabilityImplicationsSupervised2022}. Given that most of these methods, as well as most face recognition models, are CNN-based networks, research focuses on trying to interpret the results of these models using techniques such as localizable deep representations \cite{zhouLearningDeepFeatures2016} and saliency maps \cite{dabkowskiRealTimeImage2017}. Visualizing these maps as an overlay on the input image can highlight which regions in the image are most discriminative and thus have a higher impact on the network. Terhorst et al. \cite{terhorst2023pixel} introduce the concept of pixel-level face image quality and a method to assess the utility of each pixel in a face image for face recognition. This yields a pixel-level quality map that can be visualized, illustrating the utility of each pixel in the image and highlighting possible defects. If an occlusion is present on some part of the face, then the quality values of the pixels in that part of the image will be generally low compared to other pixels in the image where the face is not occluded. Fu et al. \cite{fuRelativeContributionsFacial2021} investigate the effects of selected face sub-regions (eyes, mouth, and nose) on face recognition performance. The paper studies the correlation of the general image quality of these sub-regions to the overall face image quality and the face recognition performance. The study concludes that the quality of all sub-regions is strongly correlated to FIQA and FR performance, with the eye region being the most influential. Fu and Damer \cite{fuExplainabilityImplicationsSupervised2022} propose an explainability approach that uses activation maps to visualize the responses of the face recognition network. These activation maps link the content of face embeddings produced by the FR network to pixels in the input face image, indicating which pixels have a higher impact on the network. Analyzing these maps for face images of different quality enables the analysis of the response of FR models to low- and high-quality face images. Visualizing these maps as overlays over the original image is then used as an interpretation of the quality of the face image.

\subsection{Retrieval Augmented Generation (RAG)}
\label{sec:rag}

LLMs have come a long way in the past few years, despite still having challenges such as hallucinations, outdated knowledge, and non-transparent, untraceable reasoning processes \cite{chen2024benchmarking, chang2024survey}. Hallucinations happen when the LLM receives queries extending beyond its training data or events that took place after the point in time the model was trained, and so the model comes up with imagined answers \cite{ji2023survey}. Retrieval Augmented Generation (RAG), introduced by Lewis et al. \cite{lewisRetrievalAugmentedGenerationKnowledgeIntensive2020}, tries to address this particular problem. RAG mitigates the problem of potentially producing factually inaccurate information by retrieving pertinent documents from external knowledge sources while still utilizing the powerful conversational capabilities of the LLM \cite{lewisRetrievalAugmentedGenerationKnowledgeIntensive2020}. RAG proposes a framework that combines parametric memory (learning-based from the training process) with non-parametric memories (retrieval-based from external data sources), making it more suitable for knowledge-intensive natural language processing (NLP) tasks \cite{cuconasu2024power}. The original RAG approach combines a pre-trained retriever for encoding queries and indexing retrieved documents with a pre-trained seq2seq model (generator) and an end-to-end fine-tuning process \cite{lewisRetrievalAugmentedGenerationKnowledgeIntensive2020}. RAG technology has rapidly developed with many variations and extensions over the original approach, though the primary RAG process still involves three main processes, namely, indexing, retrieval, and generation \cite{ji2023survey, cai2022recent}.

\subsection{LLM-Powered Autonomous Agents}
\label{sec:agent}

The basic functionality of an LLM is that, given an input \textit{prompt}, it outputs an answer using the knowledge it internalized during the training process \cite{chang2024survey, yao2024survey}. Augmentation techniques, such as RAG, augment this process by giving the LLM access to external sources of knowledge, such as private databases and document stores \cite{lewisRetrievalAugmentedGenerationKnowledgeIntensive2020}. In an LLM-powered autonomous agent system, the LLM functions as the agent's brain, utilizing three key concepts: planning, memory, and tools \cite{wang2024survey}. Planning is necessary for the agent to reason about complicated tasks and break them down into smaller, manageable sub-goals. Prompting techniques such as \textit{Chain of Thoughts (CoT)} and \textit{Tree of Thoughts (ToT)} are common for this purpose \cite{yao2024tree}. In the CoT, the LLM is instructed through a chain of prompts to think in a step-by-step approach, decomposing the task into smaller and simpler steps \cite{wei2022chain}. The ToT extends CoT by exploring multiple reasoning possibilities at each step, creating a tree structure \cite{yao2023tree}. Memory, utilized by an agent, can be thought of in terms of the three types of memory in the human brain. These are: (1) sensory memory, which gives access to sensory information such as visual and auditory signals; (2) short-term or working memory, which stores the information we are currently aware of; and (3) long-term memory, which stores information for a long time, including facts, events, and skills and routines that are performed automatically \cite{BREWER19831, wang2024survey}. Tool usage, also inspired by human behavior, allows the agent to extend its capabilities and utilize external tools to achieve the tasks it has reasoned about. Many works in the literature \cite{shen2024hugginggpt, schick2024toolformer}, as well as commercial tools such as ChatGPT \textit{Function calling}\footnote{https://platform.openai.com/docs/guides/function-calling}, have used this concept to give the LLM access to external APIs that can perform specialized tasks.

%% file: 03_approach.tex
\section{Proposed Method}
\label{sec:approach}

\begin{figure}[h]
     \centering
     \includegraphics[width=\textwidth]{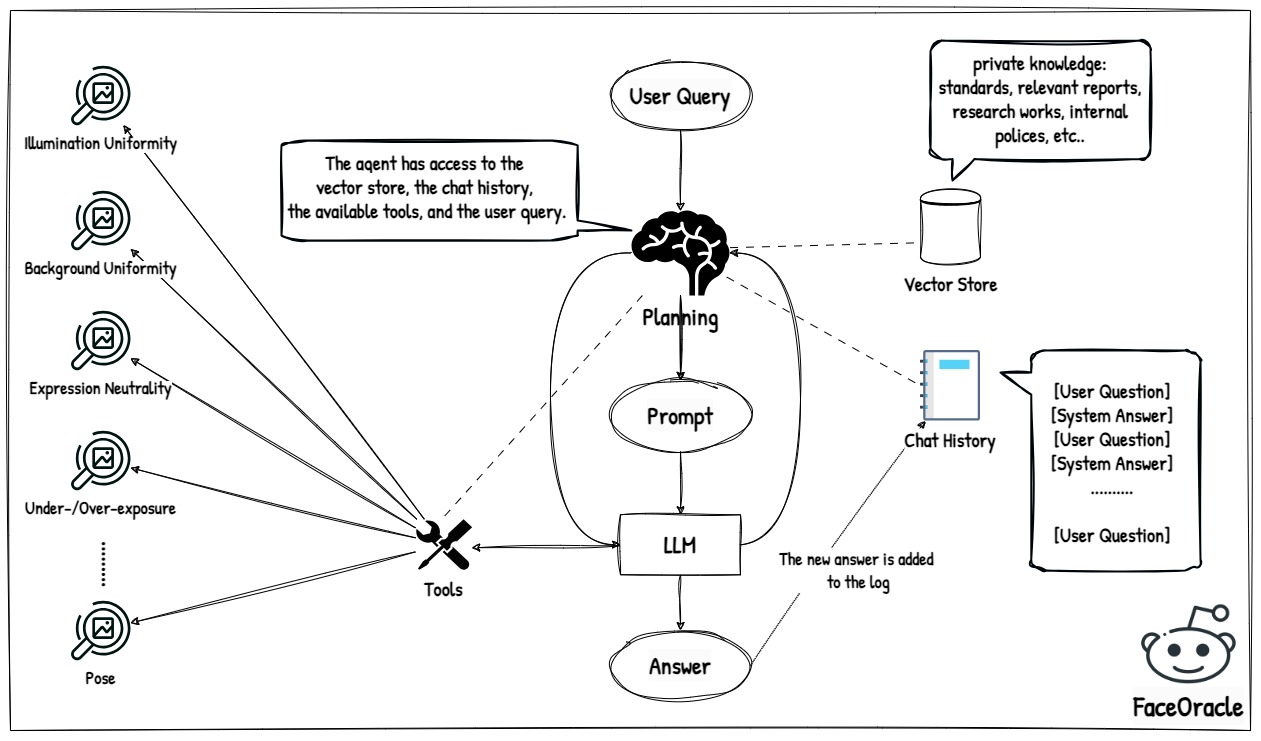}
    \caption{The overall scheme of FaceOracle. FaceOracle integrates (1) a set of tools (FIQA algorithms), (2) private knowledge sources (standards, literature, internal policies), and (3) a chat history to enable coherent, consistent, and context-aware conversations. A user at passport issuing authority office submits a query to FaceOracle, which reasons about the query, makes any necessary computations (e.g., computes the quality score of an image), retrieves any relevant data from the knowledge sources, takes into consideration the chat history, and finally uses the power of LLM to formulate the final answer in a natural language to the user.}
    \label{fig:scheme}
\end{figure}

This section explains the proposed scheme for the face image analysis AI assistant, FaceOracle. The scheme uses the Retrieval Augmented Generation (RAG) technique introduced in \cref{sec:rag} and the main ideas of LLM-powered autonomous agents in \cref{sec:agent}. \cref{sec:metho-rag} describes how the RAG technique is used to give the agent access to needed domain knowledge, while \cref{sec:metho-agent} describes how FaceOracle is constructed as an LLM-powered autonomous agent with access to domain knowledge and specialized tools. The proposed scheme is illustrated in \cref{fig:scheme}.

\subsection{FaceOracle Domain Knowledge}
\label{sec:metho-rag}

FIQA \cite{ISO-IEC-29794-5-DIS-FaceQuality-240129} is a specialized domain that validates the compliance of face images with detailed requirements described in several international standards \cite{ISO-IEC-19794-5-G2-FaceImage-110304, ISO-IEC-39794-5-G3-FaceImage-191015}. Available LLMs are general-purpose and trained over publicly available data; thus, it is not expected that they give precise answers in such specialized domains \cite{harvel2024can}. To create an AI assistant that can answer questions about a specialized field such as FIQA and can give correct, not-hallucinated answers grounded in the standards and other relevant documents, we use RAG to give FaceOracle access to FIQA standards, relevant literature works, and sample internal face image quality policies of an issuing authority. This empowers FaceOracle to provide users with precise and reliable feedback, customized to their specific organization. The RAG process, as suggested in FaceOracle, consists of the following three stages:

\subsubsection{Indexing:} This is the process of indexing the private external data sources, including the documents of the standards, internal policies, and relevant literature. We divide the text of these documents into smaller pieces, known as chunks. We convert the chunks into embeddings, a vector representation of the text that internalizes semantic information, and store them in a vector store, which serves as the external knowledge source.

\subsubsection{Retrieval:} The questions asked by the user (queries) will be embedded using the same embedding process used to embed the text of the external data sources to make sure both are embedded into the same high-dimensional space. Given that the embedding vectors contain semantic information, we can perform a semantic search in the vector store, retrieve the relevant embeddings, and thus the pieces of text that are most relevant for the query, and then rank the results.

\subsubsection{Generation:} The retrieved ranked results from the retrieval step will be used as part of the prompt sent to the LLM. This will provide the \textit{context} and the source of information on which the generated answers should be based.

\subsection{FaceOracle Agent}
\label{sec:metho-agent}

The FaceOracle agent utilizes the three key concepts of planning, memory, and tool usage, with the powerful capabilities of a pre-trained LLM, to analyze, reason, and respond to user queries.

\subsubsection{Planning:} As discussed in \cref{sec:agent}, agents use planning to understand and reason about their tasks. As opposed to simply sending the raw user query directly to the LLM and receiving an answer, FaceOracle interacts with the LLM in multiple cycles before generating the final answer for the user. The agent first needs to determine whether it needs to call any of the tools (FIQA algorithms) or query the vector store (which possesses specialized domain knowledge about FIQA) to fulfill the user query, necessitating this multi-cycle process. Once it gathers responses from the tools and vector store, if necessary, it forwards these responses, along with the chat history and the user query, to the LLM. The LLM then generates the final answer and delivers it to the user.

\subsubsection{Memory:} FaceOracle is designed to make use of the three types of memory described in \cref{sec:agent}. First, FaceOracle uses a pre-trained LLM; thus, the internalized knowledge in the model is considered the first form of long-term memory that the agent has access to. We further augment this, using the concept of RAG, and give the agent access to domain knowledge in the form of non-publicly available standardization documents and the issuing authority's policies. This prevents the agent from hallucinating answers and ensures that it gives trusted answers based on standards and policies. Next, the agent realizes the working memory concept by gaining access to the chat history. Without it, the agent would treat each query as an entirely independent input, disregarding past interactions. Lastly, the immediate user query that initiates the flow is considered to serve as sensory memory.

\subsubsection{Tools:} The most crucial component the agent needs is access to the specialized FIQA algorithms to be able to perform various FIQA operations needed to fulfill incoming queries. Therefore, the agent connects to a set of FIQA algorithms through the concept of \textit{tools}, enabling it to obtain relevant quality assessments.

%% file: 04_experiments.tex
\section{Implementation}
\label{sec:implementation}

To realize the scheme described in \cref{sec:approach} and illustrated in \cref{fig:scheme}, a set of tools needs to be selected and used to implement FaceOracle. This section outlines the types of tools required, as well as the specific tools selected for the proof-of-concept created as part of this work.

\subsubsection{Vector Store:} A vector store is a specialized type of database optimized for storing, indexing, and querying vector data \cite{SubramaniamVS2006}. Vector data refers to the high-dimensional numerical representations (embeddings) of the text chunks from external knowledge sources that we want to incorporate in FaceOracle. A vector store has the ability to perform similarity searches, which involve finding vectors that are close to a given query based on a distance measure such as Euclidean distance, cosine similarity, or other mathematical definitions of "closeness." We chose to use \textit{chroma} because it is an open-source and free vector store that allows efficient similarity search\footnote{https://www.trychroma.com/}.

\subsubsection{Embedding Model:} We need a pre-trained embedding model to convert the text chunks retrieved from the documents into embeddings. We chose to use \textit{Instructor} because it is a general embedder model that creates text embeddings that are suitable for different tasks and domains without any extra training \cite{su-etal-2023-one}. Moreover, it is free and still performs better than other paid options, such as \textit{text-embedding-ada-002}\footnote{https://openai.com/index/new-and-improved-embedding-model/} from OpenAI \cite{muennighoff-etal-2023-mteb}.

\subsubsection{LLM:} We need a powerful LLM to serve as the system's brain and facilitate conversational interactions with the user. Many open-source LLMs are available, such as Llama 2 \cite{touvron2023llama}, Falcon \cite{penedo2023refinedweb}, and Mistral \cite{jiang2023mistral}, but for ease of setup, we use ChatGPT, specifically \textit{gpt-3.5-turbo}\footnote{https://openai.com/index/chatgpt/}, for the proof-of-concept. However, to ensure the confidentiality of private data in a real setup, we recommend using another open-source or privately deployed LLM.

\subsubsection{Application Framework:} We use LangChain, a framework for developing applications powered by LLMs)\footnote{https://www.langchain.com/}, to integrate the LLM with external knowledge sources as well as the tools.

\subsubsection{FIQA Algorithms:} We need a set of FIQA algorithms to analyze the face image and extract quality component values. To this end, we use the Open Source Face Image Quality (OFIQ)\footnote{https://github.com/BSI-OFIQ/OFIQ-Project} framework. OFIQ is the reference implementation for the ISO/IEC 29794-5 standard \cite{ISO-IEC-29794-5-DIS-FaceQuality-240129} which guarantees that FaceOracle has access to a set of fully compliant FIQA algorithms.

%% file: 05_discussion.tex
\section{Evaluation}
\label{sec:evaluation}

\subsection{Quantitative Evaluation}

Since the concept of LLM-powered agents is new to the field of face image quality assessment (FIQA), we could not find other works in the literature that form baselines, introduce evaluation datasets, or propose evaluation criteria or metrics. Therefore, we present evaluation criteria inspired by other works on LLM-powered agents \cite{liu2023agentbench, chan2023chateval}. We propose a set of metrics to quantitatively evaluate how well our agent performs on this criteria. We introduce an evaluation dataset and compare the results, when possible, to ChatGPT.

\subsubsection{Evaluation Criteria}

The evaluation criteria we use for FaceOracle should address the most critical requirements the system should possess in order to be useful in a real operational setting. Specifically, we evaluate the following:

\begin{itemize}
    \item Correctness: Does the agent's response align with an established ground truth answer? This means that the information, such as definitions or explanations, in the answer is correct and matches expected ground truth information, and that any computation outputs, such as the face image quality assessment values of an image, also match expected ground truth values.
    \item Relevance: Are the documents the agent retrieves from the vector store relevant to the user's question? This is to ensure that FaceOracle retrieves the correct domain knowledge and provides context to use in formulating the user's answer.
    \item Faithfulness: Is the answer grounded in the retrieved documents? This ensures that FaceOracle does not hallucinate the answer.
\end{itemize}

An evaluation criteria that includes these three elements will provide us with information on whether the agent is retrieving relevant documents in response to a user's query. Once the agent has retrieved the relevant documents, the next step is to determine whether the output answer is based on these documents and not a result of hallucinations. Finally, determine whether the information and any computed values in the final answer are correct.

\subsubsection{Evaluation Dataset}

As mentioned earlier, there is no available dataset for evaluation, so we introduce one. Given that the questions that FaceOracle might encounter are of two types: those that require using FIQA algorithms to compute quality measures and other general questions about FIQA concepts that do not require the use of FIQA algorithms, we created an evaluation dataset with two types of samples.

For the first type of sample, we use face images from the FRLL dataset \cite{DeBruine2021}. We compute their ground truth quality values using OFIQ, formulate questions about one or several quality measures for each image, and provide the expected ground truth answer. We create a total of 1000 samples of this type. \cref{tab:subsetA} shows an example of this type of sample. For the second sample type, we formulate questions about FIQA concepts and also provide the expected ground truth answer. We create a total of 100 samples of this type. \cref{tab:subsetB} shows an example of this type of sample.

\begin{table}
\centering
\caption{Each sample comprises: (1) an image name referring to the face image; (2) a question; and (3) the ground truth answer.}\label{tab:subsetA}
\begin{tabular}{|p{0.2\linewidth} | p{0.7\linewidth}|}
\hline
Key & Value \\
\hline
Image & sub1-001.jpg \\
\hline
Question & what are the expression neutrality and illumination uniformity quality values of this image?  \\
\hline
Answer & The expression neutrality measure has the value of 70, and illumination uniformity has the value of 59. \\
\hline
\end{tabular}
\end{table}

\begin{table}
\centering
\caption{Each sample comprises: (1) a question; and (2) the ground truth answer.}\label{tab:subsetB}
\begin{tabular}{|p{0.2\linewidth} | p{0.7\linewidth}|}
\hline
Key & Value \\
\hline
Question & What is unified quality score?  \\
\hline
Answer & A quantitative expression of the predicted verification performance of the biometric sample. Valid values for Quality Score are integers between 0 and 100, where higher values indicate better quality. \\
\hline
\end{tabular}
\end{table}

\subsubsection{Evaluation Metrics and Results}

To quantitatively measure how well the agent is performing on the presented criteria, We propose the following four metrics:

\begin{itemize}
    
    \item Tool Selection Accuracy (TSA): the number of times the agent selects the correct tool (FIQA algorithm) to perform the computation over the total number of expected tool calls.

    \item Average Answer Reference Distance (ARD): The average distance between the agent's answer and the reference ground truth answer across all evaluation samples.
    
    \item Average Answer Context Distance (ACD): The average distance between the agent's answer and the context (the retrieved documents used to formulate the answer) across all evaluation samples.
    
    \item Average Question Context Distance (QCD): The average distance between the user's question and the context across all evaluation samples.
    
\end{itemize}

The \textit{distance} here refers to the cosine distance between two vectors after converting each of the two strings relevant to each metric to an embedding vector. We chose to use TSA instead of, for example, Mean Squared Error (MSE) because, in our case, selecting the right tool, such as the algorithm for assessing neutral expression when the question indicates it needs this measure, will guarantee that the computed value is the same as the ground truth since both are computed using the same exact algorithm from OFIQ. MSE or similar metrics could replace this when evaluating other systems using this dataset. The first two metrics, TSA and ARD, together measure the correctness of the answers. Therefore, the higher the TSA and the smaller the ARD, the higher the correctness of the system. The third metric measures the faithfulness of the system. The smaller the ACD, the more grounded the answer in context, and the less hallucinated it is by the LLM. The fourth metric measures the relevance of the context to the question. Smaller QCD values mean that the agent retrieves relevant documents for the question from the vector store and uses them as context to formulate the answer.

TSA is only relevant for samples of type 1, while ACD, QCD, and ARD are only relevant for samples of type 2. For ChatGPT, we can only compute ARD because it lacks access to both the FIQA algorithms and the private knowledge source. A summary of the results is shown in \cref{tab:resutls}. The results show that FaceOracle accurately determines the appropriate tools to use based on a user's question and that it is retrieving documents relevant to the questions and formulating answers based on these documents. They also demonstrate how FaceOracle's answers are semantically closer to the questions in the evaluation set than ChatGPT's ones.

\begin{table}
\centering
\caption{Quantitative evaluation metrics. TSA is computed on type 1 samples. ACD, QCD, and ARD are computed on type 2 samples, with only ARD having results for both ChatGPT and FaceOracle.}
\label{tab:resutls}
\begin{tabular}{|l|l|l|l|l|}
\hline
Method & TSA & ACD & QCD & ARD \\
\hline
ChatGPT & - & - & - & 0.23 \\
FaceOracle & 0.97 & 0.12 & 0.17 & 0.081 \\
\hline
\end{tabular}
\end{table}

\subsection{Qualitative Evaluation}

\begin{figure}
     \begin{center}
     \begin{subfigure}[b]{0.49\linewidth}
         \centering
         \includegraphics[width=\linewidth]{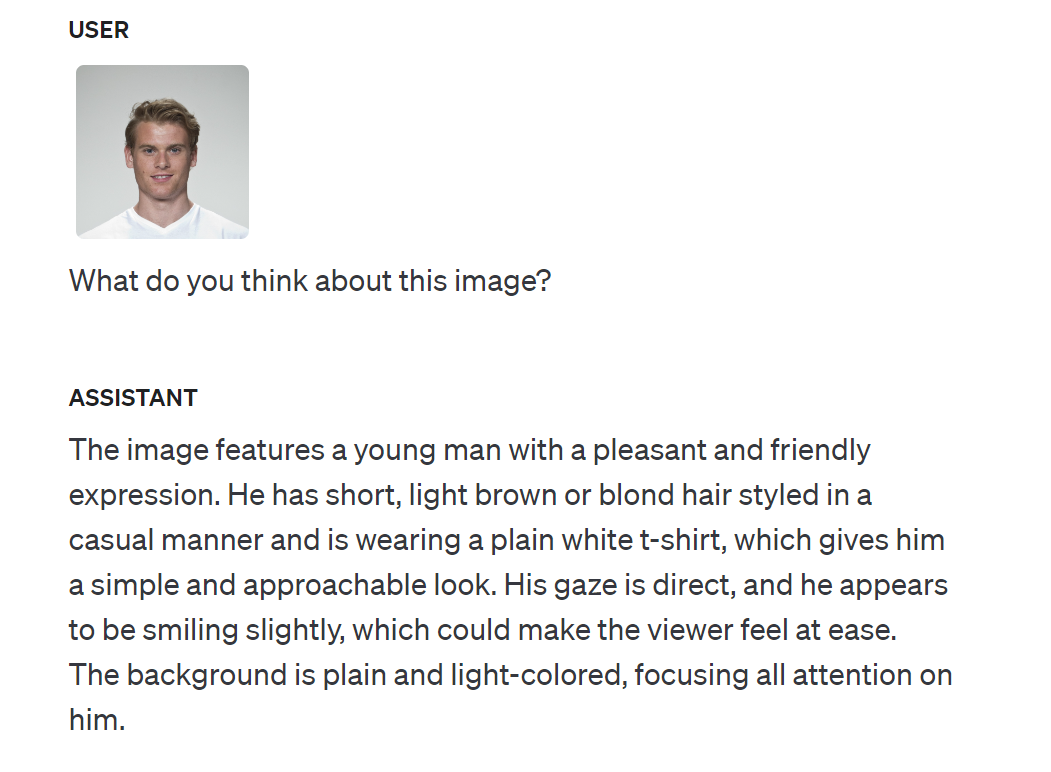}
         \caption{ChatGPT}
         \label{fig:wdut-gpt}
     \end{subfigure}
    \hfill
     \begin{subfigure}[b]{0.49\linewidth}
         \centering
         \includegraphics[width=\linewidth]{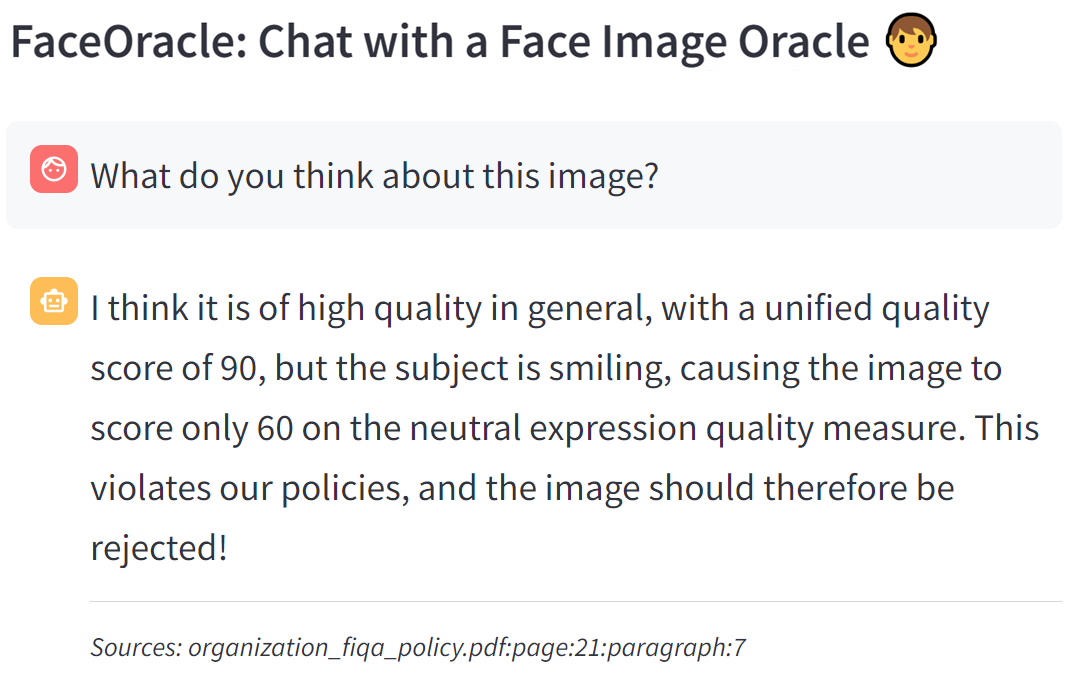}
         \caption{FaceOracle}
         \label{fig:wdut-face}
     \end{subfigure}
     \end{center}
    \caption{Comparing two actual answers from ChatGPT and FaceOracle to the same generic question about an image. ChatGPT gives a very generic description of the image, while FaceOracle gives a unified quality score, points to a defect, and provides a suggested action. Face image from the FRLL dataset \cite{DeBruine2021}.}
    \label{fig:wdut}
\end{figure}

\begin{figure}
     \begin{center}
     \begin{subfigure}[b]{0.49\linewidth}
         \centering
         \includegraphics[width=\linewidth]{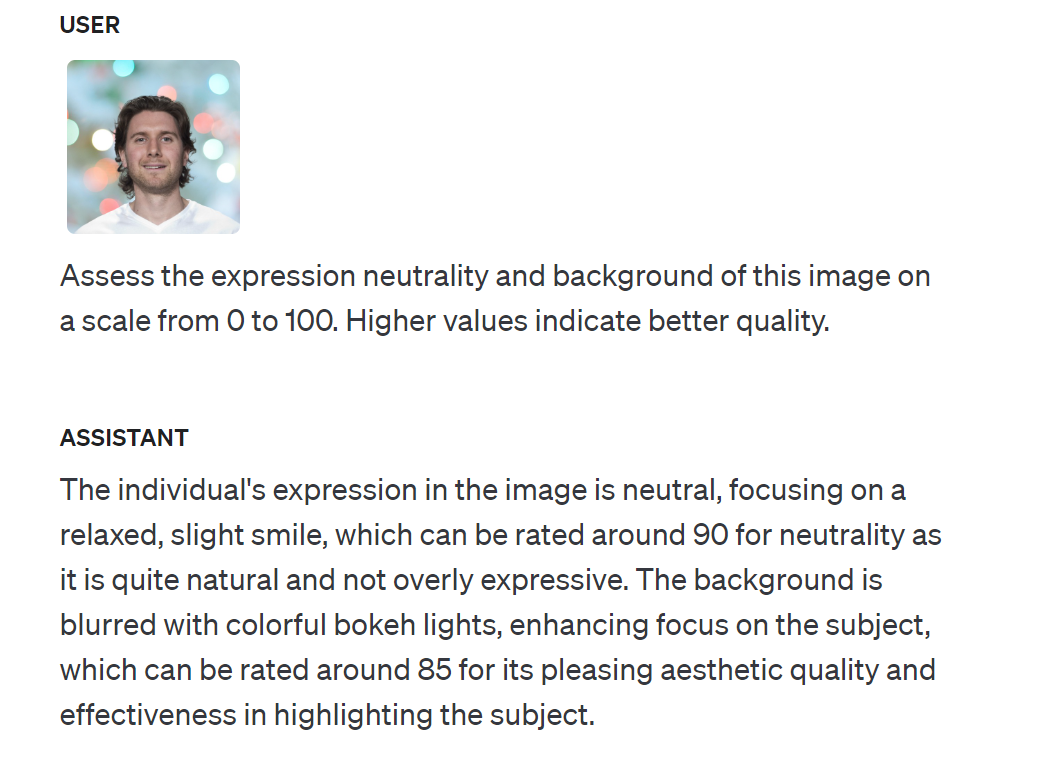}
         \caption{ChatGPT}
         \label{fig:enbg-gpt}
     \end{subfigure}
    \hfill
     \begin{subfigure}[b]{0.49\linewidth}
         \centering
         \includegraphics[width=\linewidth]{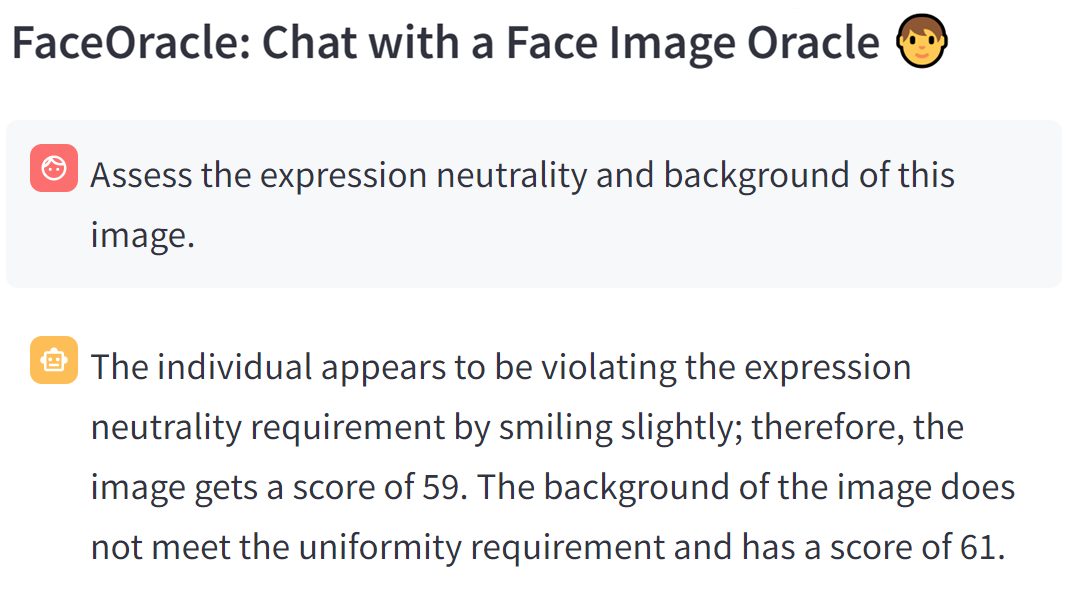}
         \caption{FaceOracle}
         \label{fig:enbg-face}
     \end{subfigure}
     \end{center}
    \caption{An example where we instruct ChatGPT to assign a numerical value to two quality measures. It performs a generic assessment and determines that the subject has a natural, acceptable smile and a nice, colorful background, resulting in high quality values for both measures. However, according to ICAO specifications, the non-uniform background and smile are not acceptable, which is why the FIQA algorithms reflect this with lower values in FaceOracle. Face image from the FRLL dataset \cite{DeBruine2021}.}
    \label{fig:enbg}
\end{figure}

\begin{figure}
     \begin{center}
     \begin{subfigure}[b]{0.49\linewidth}
         \centering
         \includegraphics[width=\linewidth]{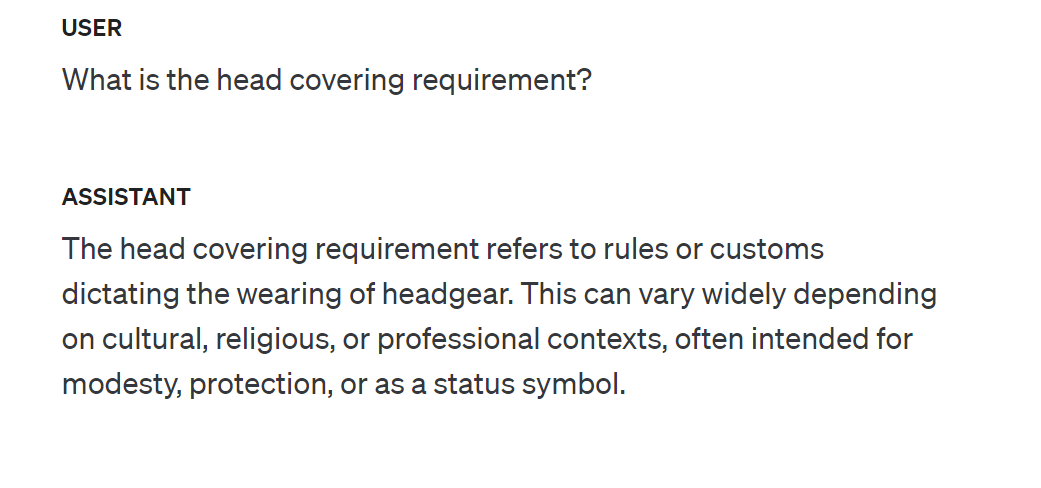}
         \caption{ChatGPT}
         \label{fig:head-covering-gpt}
     \end{subfigure}
    \hfill
     \begin{subfigure}[b]{0.49\linewidth}
         \centering
         \includegraphics[width=\linewidth]{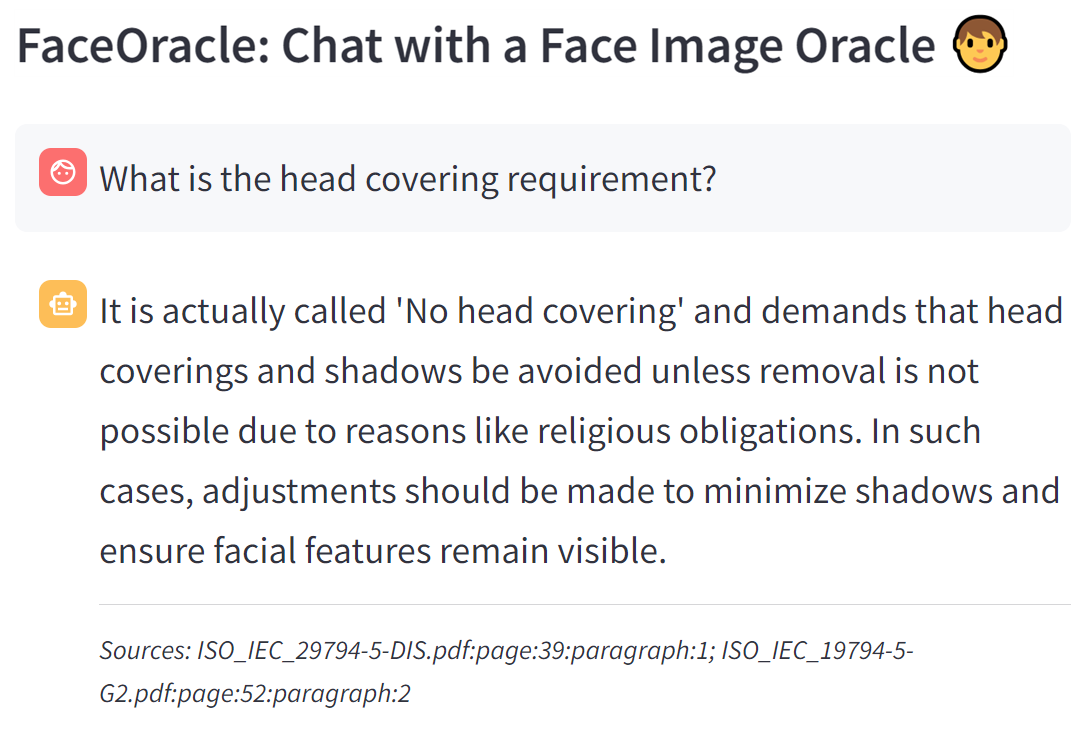}
         \caption{FaceOracle}
         \label{fig:head-covering-face}
     \end{subfigure}
     \end{center}
    \caption{Comparing two actual answers from ChatGPT and FaceOracle to the same question about head covering. ChatGPT is giving a very generic answer about head coverings, while FaceOracle is giving a face image quality related answer with references to the standards.}
    \label{fig:head-covering}
\end{figure}

\begin{figure}
     \begin{center}
     \begin{subfigure}[b]{0.49\linewidth}
         \centering
         \includegraphics[width=\linewidth]{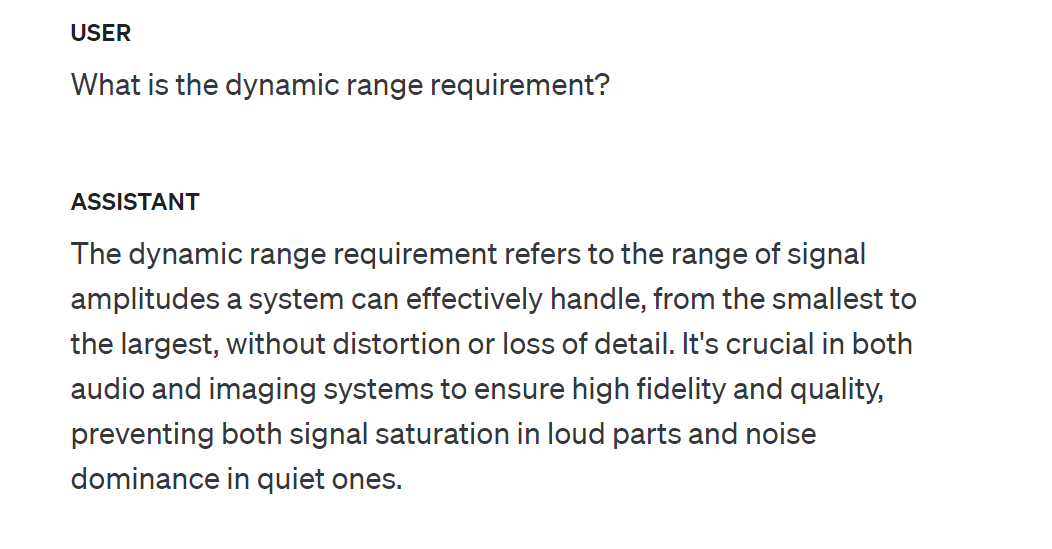}
         \caption{ChatGPT}
         \label{fig:dr-gpt}
     \end{subfigure}
    \hfill
     \begin{subfigure}[b]{0.49\linewidth}
         \centering
         \includegraphics[width=\linewidth]{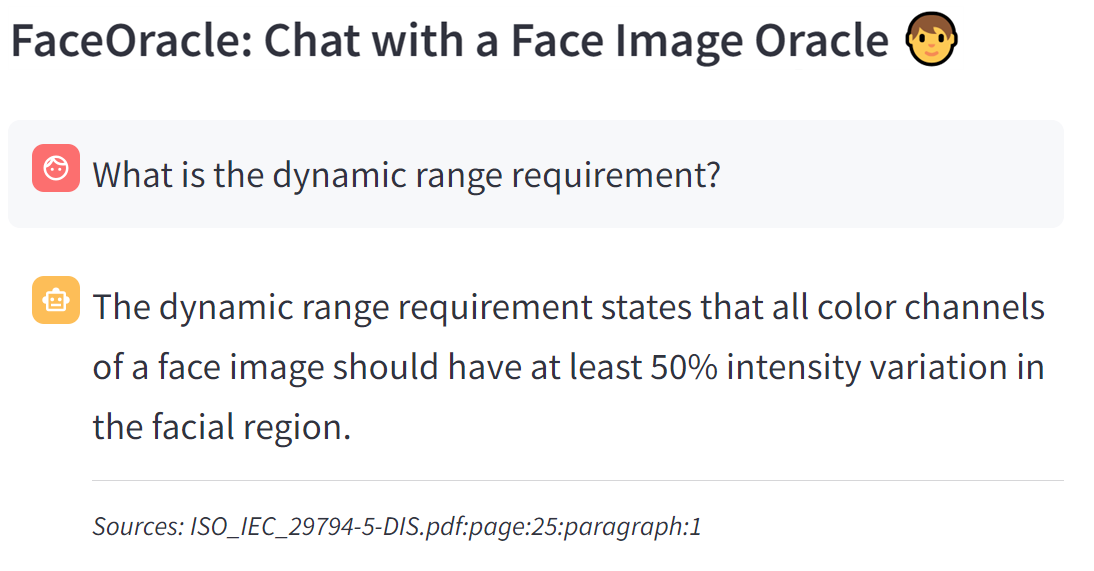}
         \caption{FaceOracle}
         \label{fig:dr-face}
     \end{subfigure}
     \end{center}
    \caption{Comparing two actual answers from ChatGPT and FaceOracle to the same question about dynamic range. ChatGPT is giving a very generic answer about dynamic range, while FaceOracle is giving a face image quality-related answer with references to the standards.}
    \label{fig:dr}
\end{figure}

Sample conversations comparing FaceOracle to ChatGPT\footnote{\textit{gpt-4-turbo}, https://platform.openai.com/docs/models} are shown in \cref{fig:wdut,fig:enbg,fig:head-covering,fig:dr}. These examples clearly illustrate the unique advantages FaceOracle has as an LLM-powered agent specialized in the FIQA domain, with access to the standardized FIQA algorithms (tools) and domain knowledge (standards and internal policies) over the generic ChatGPT, which certainly, being a general purpose LLM, does not have access to such capabilities.

\begin{figure}
    \centering
    \includegraphics[width=\textwidth]{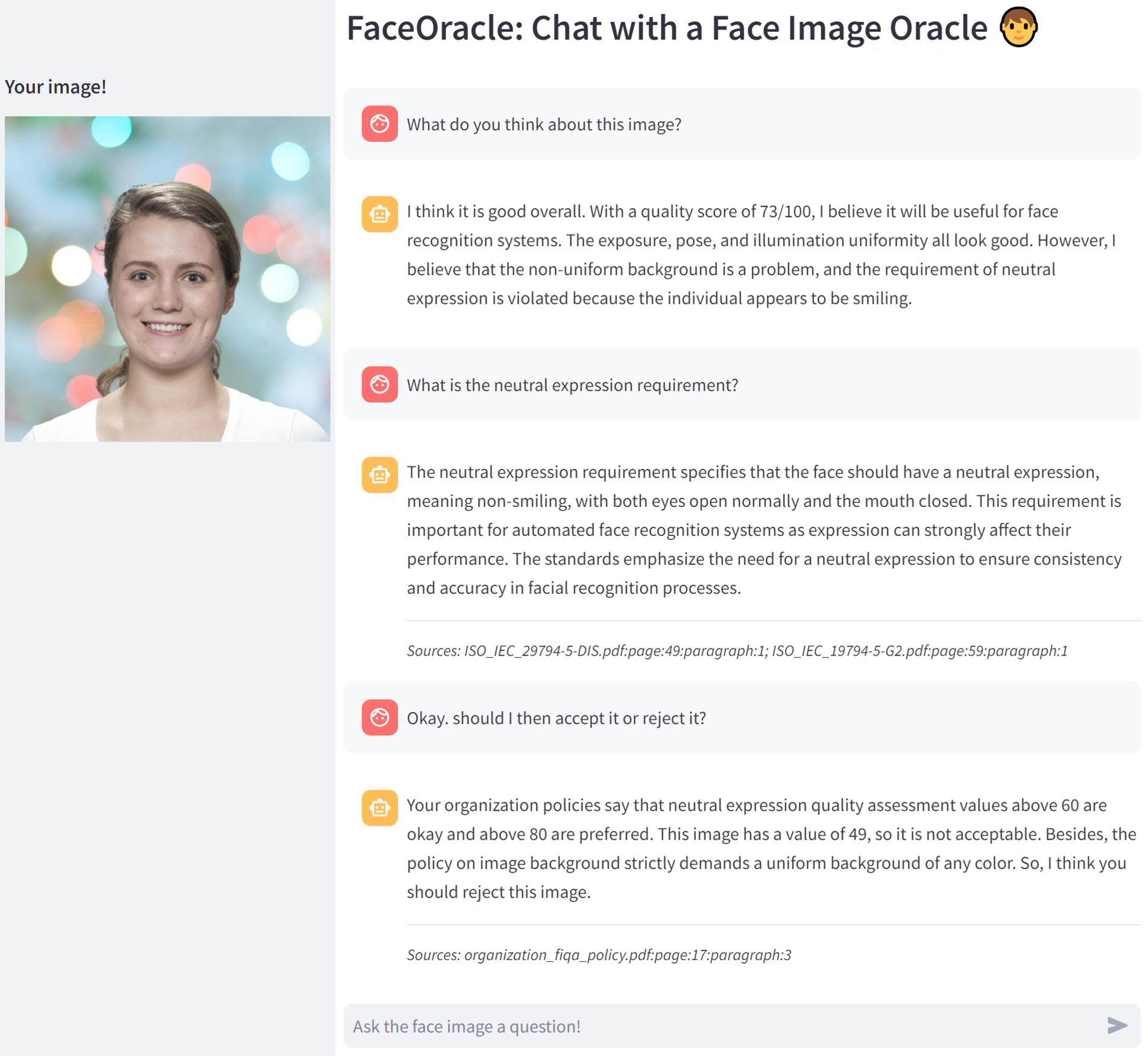}
    \caption{A sample chat with FaceOracle, in which FaceOracle invokes various FIQA algorithms and queries and retrieves information from the standards and the issuing authority's policies to help the user perform a quality assessment of the image, understand the results, and make a decision, FaceOracle clearly states the source documents from which it extracts information, including the document name, page, and exact paragraph, thereby assisting the user in justifying and communicating their decisions. Face image from the FRLL dataset \cite{DeBruine2021}.}
    \label{fig:chat}
\end{figure}

%% file: 06_conclusion.tex
\section{Conclusion}
\label{sec:conclusion}

In this work, we introduced FaceOracle\footnote{https://github.com/wkabbani/FaceOracle}, a new approach to explainability in face image quality assessment and biometrics in general that puts the user experience at the forefront. It describes a unique scheme of concepts, techniques, and tools that together realize an AI assistant that can utilize standardized FIQA algorithms, access private and specialized domain knowledge, and help its users analyze, understand, decide, and communicate results and decisions. To evaluate FaceOracle, we introduced a dataset, formalized assessment criteria, and established a set of metrics for quantitative performance evaluation.

FaceOracle is intended to serve as an AI assistant for various tasks associated with face image analysis, including FIQA, face image manipulation detection, and morphing detection. This work realized FaceOracle for FIQA, but utilizing the presented concepts, these extra capabilities can be easily added into FaceOracle, which could be the purpose of future work.

%% file: 10_other.tex
\section*{Acknowledgment}
\label{sec:acknowledgment}

This work was supported by the European Union's Horizon 2020 Research and Innovation Program under Grant 883356.